\documentclass[letterpaper]{article} % DO NOT CHANGE THIS
\usepackage{aaai25}  % DO NOT CHANGE THIS
\usepackage{times}  % DO NOT CHANGE THIS
\usepackage{helvet}  % DO NOT CHANGE THIS
\usepackage{courier}  % DO NOT CHANGE THIS
\usepackage[hyphens]{url}  % DO NOT CHANGE THIS
\usepackage{graphicx} % DO NOT CHANGE THIS
\usepackage{natbib}  % DO NOT CHANGE THIS AND DO NOT ADD ANY OPTIONS TO IT
\usepackage{caption} % DO NOT CHANGE THIS AND DO NOT ADD ANY OPTIONS TO IT
\usepackage{algorithm}
\usepackage{algorithmic}

\usepackage{newfloat}
\usepackage{listings}
\DeclareCaptionStyle{ruled}{labelfont=normalfont,labelsep=colon,strut=off} % DO NOT CHANGE THIS
\floatstyle{ruled}
\newfloat{listing}{tb}{lst}{}
\floatname{listing}{Listing}
\usepackage{amsmath}
\usepackage{amssymb}
\usepackage{tabularray}
\usepackage{multirow}
\usepackage{subcaption}
\usepackage{booktabs}

\usepackage{amsthm}
\newtheorem{definition}{Definition}

\newtheorem{theorem}{Theorem}

\title{LOHA: Direct Graph Spectral \\Contrastive Learning Between Low-pass and High-pass Views}
\author{
    Ziyun Zou\textsuperscript{\rm 1},
    Yinghui Jiang\textsuperscript{\rm 2},
    Lian Shen\textsuperscript{\rm 1},
    Juan Liu\textsuperscript{\rm 3},
    Xiangrong Liu\textsuperscript{\rm 1,2}\thanks{Corresponding Author}\\
}
\affiliations {
    \textsuperscript{\rm 1}Department of Computer Science and Technology, Xiamen University,\\
    \textsuperscript{\rm 2} National Institute for Data Science in Health and Medicine, Xiamen University,\\
    \textsuperscript{\rm 3}Pen-Tung Sah Institute of Micro-Nano Science and Technology, Xiamen University,\\
    \{zoezzy, shenlian\}@stu.xmu.edu.cn, stardjyeah@gmail.com,
    \{cecyliu, xrliu\}@xmu.edu.cn
}

\begin{document}

\maketitle

\begin{abstract}
Spectral Graph Neural Networks effectively handle graphs with different homophily levels, with low-pass filter mining feature smoothness and high-pass filter capturing differences. When these distinct filters could naturally form two opposite views for self-supervised learning, the commonalities between the counterparts for the same node remain unexplored, leading to suboptimal performance. In this paper, a simple yet effective self-supervised contrastive framework, LOHA, is proposed to address this gap. LOHA optimally leverages low-pass and high-pass views by embracing "harmony in diversity". Rather than solely maximizing the difference between these distinct views, which may lead to feature separation, LOHA harmonizes the diversity by treating the propagation of graph signals from both views as a composite feature. Specifically, a novel high-dimensional feature named spectral signal trend is proposed to serve as the basis for the composite feature, which remains relatively unaffected by changing filters and focuses solely on original feature differences. LOHA achieves an average performance improvement of 2.8\% over runner-up models on 9 real-world datasets with varying homophily levels. Notably, LOHA even surpasses fully-supervised models on several datasets, which underscores the potential of LOHA in advancing the efficacy of spectral GNNs for diverse graph structures. 
\end{abstract}

\section{Introduction}

Graphs have been widely used to model the relationships between entities. In homophilic graphs, connected nodes tend to share similar features; such as in citation networks~\cite{GCN}. Conversely, in heterophilic graphs, the opposite is true, as seen in dating networks~\cite{evennet}. Mining graph information has proven immensely helpful in fields such as bioinformatics~\cite{msgcl,molecular}, link prediction~\cite{link_nips,lu_link}, recommendation systems~\cite{recommendation, kdd_recommend}, and traffic prediction~\cite{traffic}. However, traditional graph neural networks(such as GCN) adopt a low-pass filter encoder to smooth the representations of neighboring nodes, limiting the adaptability to heterophilic graphs~\cite{h2gcn,all_we_have}. 

Thus, more explorations towards designing delicate Spectral Graph Neural Networks(SGNNs) are conducted~\cite{chebnet,how_powerful,slog}. Being benefited from the ability to fit both low-pass and high-pass signals, it is determined as an effective solution to deal with graphs with different levels of homophily~\cite{bernnet,chebnetii}. When low-pass filters amplify the low frequencies and mine feature smoothnesses, high-pass filters are considered preferable for heterophilic graphs and could capture feature differences~\cite{fagcn,xu_heat_kernel,GAME}. In supervised learning tasks, SGNNs with polynomial approximation excel in both homophilic and heterophilic settings by adaptively fitting graph filters of arbitrary shapes~\cite{aswt-sgnn,optbasis}. Yet, for real-world applications, node labels and homophily levels of graphs are mostly uncovered. It remains necessary but challenging to learn appropriate filters for different graphs without label supervision automatically~\cite{polygcl,s3gcl}. 

In self-supervised graph representation learning, graph contrastive learning (GCL) has achieved significantly superior performance~\cite{non_positive,sfa,joao}. GRACE~\cite{grace} correlates graph views by pushing closer representations of the same node in different views and pushing apart representations of different nodes. SPAN~\cite{span} constructs two views to maximize spectral changes, declaring it as the best way to preserve structure invariance. However, only the validations on homophilic graphs are confirmed. Thus, researchers conduct more exploitation to fit in more graphs. Equipping with the natural differences and specialties among low-pass and high-pass filters, recent researches naturally construct contradictions between the views shaped by these counterparts. Nevertheless, both PolyGCL~\cite{polygcl} and S3GCL~\cite{s3gcl} refine the models based on backpropagation between low-pass or high-pass views with the final views instead of the direct contradiction between the counterparts. We concern these approaches cannot fully exploit the characteristics of these two sets of filters and the improvements are solely gained through the complex combination of filters. However, if we treat low-pass and high-pass views as mutual 'negative' samples from the filter specialties perspective—contrary to the traditional approach of maximizing mutual information between different views—it would cause the features of a node to diverge. This leads to our core question: \textit{what 'positive' features(commonalities) should the two counterparts share to reunite the node's features back, ensuring the identification and consistency of features for each node?}

To address the aforementioned questions, we revisit existing graph contrastive learning methods and propose a novel spectral-based diagram. To the best of our knowledge, we are the first to leverage the opposing characteristics of low-pass and high-pass filters to construct natural negative samples, thereby guiding the model to learn optimal filtering strategies and better features. While maximizing the differences between the views of the same node might cause node features to diverge, we reckon the "harmony" in 
"diversity" could be achieved by considering the propagation of graph signals as a composite feature, thereby enhancing the identification of nodes. Specifically, we introduce a novel high-dimensional feature named spectral signal trend, which describes the propagation difference of signals on the graph, to serve as a basis for the composite feature. With the extra composite feature, we propose \textbf{LOHA}, a direct graph spectral contrastive learning framework between \textbf{LO}w-pass and \textbf{H}igh-p\textbf{A}ss views. LOHA attains state-of-the-art results across various real-world datasets with different homophilic levels under a self-supervised learning framework, surpassing the second-best model by an average of 2.8\% on nine datasets. Performances on several heterophilic graph datasets even surpass full-supervised learning models, further showing the superiority of our framework and the experimental effectiveness of our theoretical discovery. Our contributions can be summarized as follows:
\begin{itemize}
    \item We propose LOHA, a simple yet effective self-supervised framework that directly contrasts low-pass and high-pass views based on their natural distinct specialties without additional data augmentations. 
    \item We highlight that directly considering low-pass and high-pass views as opposing sides could better utilize the specialties of filters but might lead to feature separation within a node. Further, we propose using the relatively stable cooperative graph signal trends as a composite feature to reunite features. Our theoretical and experimental analyses validate the effectiveness of this approach.
    \item Extensive experiments on 9 real-world datasets across different homophily levels are conducted to verify the superiority of LOHA. Additional detailed ablation studies further corroborate our theoretical results. 
\end{itemize}

\section{Preliminary}
\subsection{Notations}
Define the graph data as \(\mathcal{G} = (\mathcal{V}, \mathcal{E})\), where \(\mathcal{V}\) represents the set of \(N\) nodes, and \(\mathcal{E} \subseteq \mathcal{V} \times \mathcal{V}\) represents the edges between them. The feature matrix \(\mathbf{X} \in \mathbb{R}^{N \times F}\) contains node features \(\mathbf{x}_i\) for each node \(\mathbf{v}_i\). The adjacency matrix \(\mathbf{A} \in \mathbb{R}^{N \times N}\) has entries \(\mathbf{A}_{ij} = 1\) if \(e_{ij} \in \mathcal{E}\) and 0 otherwise. The normalized adjacency matrix is defined as \(\hat{\mathbf{A}} = \mathbf{D}^{-1/2} \mathbf{A} \mathbf{D}^{-1/2}\), where \(\mathbf{D}\) is the diagonal degree matrix with \(d_{ii} = \sum_j \mathbf{A}_{ij}\). Homophily describes the tendency of node connections. As described in~\cite{h2gcn}, edge homophily degree is defined as:
\begin{equation}
    h = \frac{\left|\left\{(v_i, v_j) \in \mathcal{E} : y_i = y_j\right\}\right|}{|\mathcal{E}|}
\end{equation}
This metric ranges from 0 to 1, where \(h\) close to 1 indicates high homophily, and vice versa. 
\subsection{Graph Spectral Filtering}
The graph Laplacian is given by \(\mathbf{L} = \mathbf{D} - \mathbf{A}\), and its normalized form is \(\tilde{\mathbf{L}} = \mathbf{I} - \hat{\mathbf{A}}\), where \(\mathbf{I}\) is the identity matrix. \(\tilde{\mathbf{L}}\) can be decomposed as \(\tilde{\mathbf{L}} = \mathbf{U} \boldsymbol{\Lambda} \mathbf{U}^\top\), with \(\boldsymbol{\Lambda} = \operatorname{diag}(\lambda_0, \ldots, \lambda_{N-1})\) being the diagonal matrix of eigenvalues and \(\mathbf{U}\) being the matrix of eigenvectors. Graph filtering is defined as \(\mathbf{U} g(\boldsymbol{\Lambda}) \mathbf{U}^\top\mathbf{X}\), where \(g(\boldsymbol{\Lambda})\) is a graph filter function. Polynomial approximations of \(g(\boldsymbol{\Lambda})\) is defined as \(\sum_{k=0}^K w_k \mathbf{L}^k\), thus using \(K\)-order truncation to fit \(g(\boldsymbol{\Lambda})\) and avoid the \(O(N^3)\) complexity of eigendecomposition.

\section{Pre-analysis}

To simplify, we use two sets of simplest low-pass filters \(\mathcal{F}_L\) and high-pass filters \(\mathcal{F}_H\) as examples:
\begin{equation}
\begin{aligned}
    &\mathcal{F}_L = \mathbf{I}+\mathbf{D}^{-1/2}\mathbf{A}\mathbf{D}^{-1/2} &= \mathbf{I}-\mathbf{L}, \\
    &\mathcal{F}_H = \mathbf{I}-\mathbf{D}^{-1/2}\mathbf{A}\mathbf{D}^{-1/2} &= \mathbf{I}+\mathbf{L}.
\end{aligned}
\label{simple_filters}
\end{equation}

Thus, the graph signals \(\mathbf{X}\) are filtered as follows:
\begin{equation}
\begin{aligned}
& \mathcal{F}_L *_G \mathbf{X}=\mathbf{U}[\mathbf{I}-\mathbf{\Lambda}] \mathbf{U}^{\top} \mathbf{X}=\mathcal{F}_L \cdot \mathbf{X}, \\
& \mathcal{F}_H *_G \mathbf{X}=\mathbf{U}[\mathbf{I}+\mathbf{\Lambda}] \mathbf{U}^{\top} \mathbf{X}=\mathcal{F}_H \cdot \mathbf{X} .
\end{aligned}    
\label{simple_filtered}
\end{equation}

As Eq.\ref{simple_filters} shows, when low-pass filters aggregate node features with the features of its neighbors, which represents the commonalities among nodes, high-pass filters focus on the differences between node features~\cite{fagcn}. This lays a solid foundation for treating the information from these two perspectives as mutually negative samples. Thus, recent research employs low-pass and high-pass filters in a self-supervised diagram, combining the encoders from two sub-views to construct the final-view encoders, thereby benefiting from both filtering techniques~\cite{polygcl,s3gcl}. However, currently these methods simply compare sub-views with final views, without the direct contraction between low- / high-pass views. Therefore, an intuitive question arises: Does the superb performance really benefit from the expectation differences between low-frequency and high-frequency information or simply from the complex combination of different filters? To answer this question, we conduct a set of simple demo experiments. 

\begin{figure}[t]
\centering
\includegraphics[width=\columnwidth]{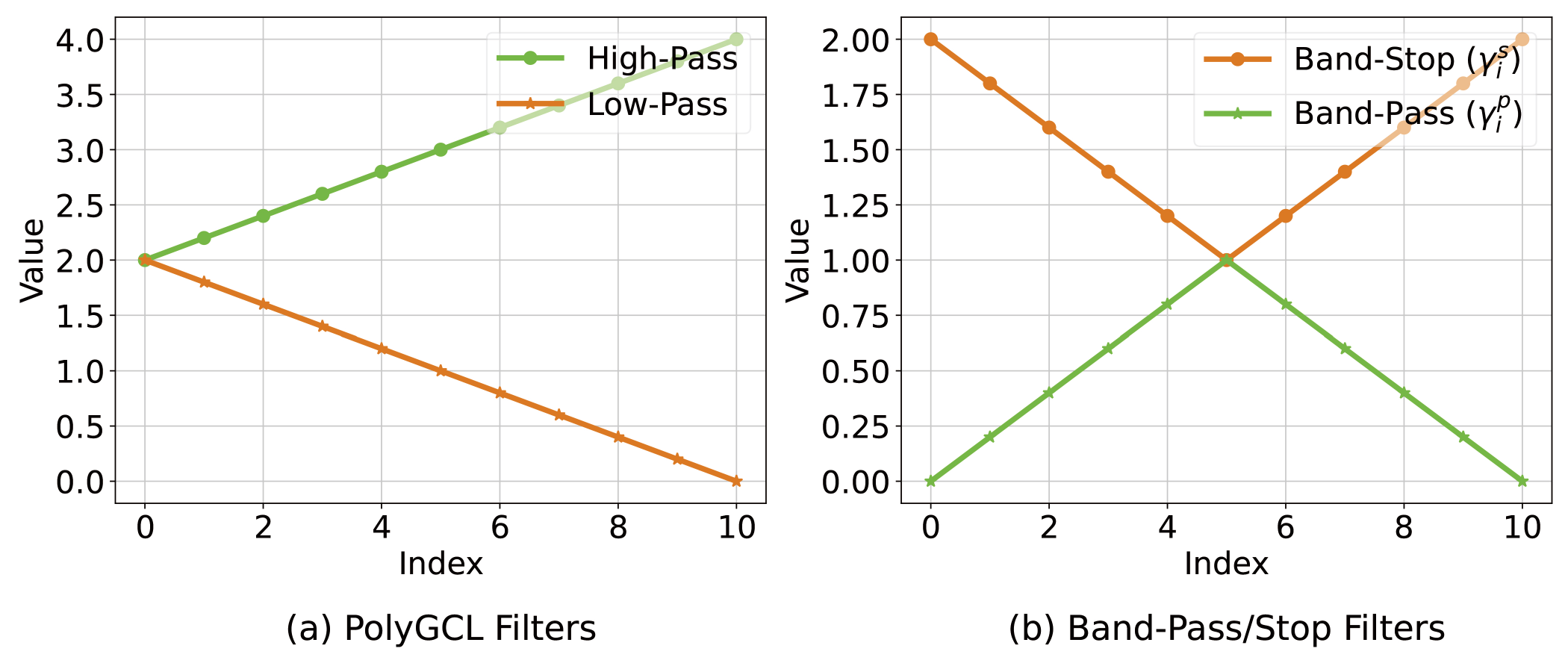}
\caption{Initialized filters for demo experiments. }
\label{fig:toy_shape}
\end{figure}

We change the low-pass and high-pass filters in PolyGCL~\shortcite{polygcl} to band-stop \(\gamma^s\) and band-pass \(\gamma^p\) sets and remain other settings unchanged to conduct comparison experiments. The filter parameters are defined as follows(also illustrated in Fig.~\ref{fig:toy_shape}(b)):

\begin{equation}
    \begin{aligned}
    &\gamma_i^s =
        \begin{cases}
            \gamma^s_0 - \sum_{j=1}^i \gamma^s_j & 1 \leq i < \left\lfloor \frac{K+1}{2} \right\rfloor \\
            \gamma^s_{K-i} & \left\lfloor \frac{K+1}{2} \right\rfloor \leq i \leq K
        \end{cases},
    \\
    &\gamma_i^p =
        \begin{cases}
            \sum_{j=0}^i \gamma^p_j \quad \quad & 1 \leq i < \left\lfloor \frac{K+1}{2} \right\rfloor \\
            \gamma^p_{K-i} \quad \quad & \left\lfloor \frac{K+1}{2} \right\rfloor \leq i \leq K
        \end{cases},
    \end{aligned}
\end{equation}
where the initial value for \(r_0^s\), \(r_K^s\) and \(r_0^p\),\(r_k^p\) are set to 2 and 0, respectively, with all the other \(\gamma\) are initialized to \(\frac{2.0}{K}\). Noted that all \(\gamma\) are trainable as in PolyGCL. Results are shown in Table~\ref{tab:toy_band}, from where we can see that band-stop and band-pass filters achieve competitive results compared with low-high filter sets, which are expected to benefit from different specialties to capture similarities or differences from features. Thus, we argue whether constructing positive and negative samples through data augmentation between sub-views and final-view instead of the natural specialties of low-high pass filters is an optimal way to benefit from these two views or not. This leads to our main question: \textit{when the direct contradiction exists in low-pass and high-pass views, what are the commonalities?} In other words, different from traditional contrastive learning like GRACE~\shortcite{grace}, which pushes representations in different views of the same node close, we propose to minimize the similarities between low-pass and high-pass views for one node to maximize the functionality of these two counterparts. However, when separating the features of one node between the low-high pass view, how to reunite features back for one node to be identifiable with other entities becomes a question.

\section{Theoretical Analysis}
We reckon that the solution is to find an extra feature to serve as a landmark during training process. It should satisfy the following two requirements: (1) Relative stability. The relative differences between nodes cannot be changed to ensure consistency. Additionally, the features should be relatively stable to ensure a smooth training process. (2) Composite. A composite feature that combines the low-pass and high-pass views as a whole to help features reunite. Inspired by \textit{Dirichlet Energy}, a global measure of how 'non-smooth' the graph information is, we propose a new basic feature to fit the first requirement. \cite{FoSR} define the \textit{Dirichlet Energy \(\mathcal{E}\)} in vector field as:

\begin{definition}[\textbf{Dirichlet Energy \(\mathcal{E}\)}]Let $\mathcal{G}$ be a connected graph with adjacency matrix \(\mathbf{A}\) and normalized Laplacian \(\mathbf{\tilde{L}}\). For \(i \in \mathcal{V}\), let \(d_i\) denote the degree of node \(i\). 
For a vector field \(\mathbf{X} \in \mathbb{R}^{n \times p}\), \(x_{i,k}\) denotes the \(k_{th}\) feature of \(\mathbf{x}_i\), we define
$$
\mathcal{E}(X):=\frac{1}{2} \sum_{i, j, k} \mathbf{A}_{i, j}\left(\frac{x_{i, k}}{\sqrt{d_i}}-\frac{x_{j, k}}{\sqrt{d_j}}\right)^2=\operatorname{Tr}\left(\mathbf{X}^T \mathbf{\tilde{L}} \mathbf{X}\right)
$$
\end{definition}
As \textit{Dirichlet energy} is a global measure, we define the \textit{Spectral Signal Trend} to describe the frequency variation of the node information on the vector field to more finely display the changing trend of one node information propagation across the node neighbors.

\begin{table}[t]
\setlength{\tabcolsep}{1mm}
\centering
\begin{tabular}{llcc}
\toprule
\multicolumn{2}{l}{Datasets}    & Low/High & Band-Pass/Stop \\
\midrule
\multirow{3}{*}{Homophilic} 
                 & Cora         &   \textbf{86.5±0.8}    &   86.3±0.9      \\
                 & Citeseer     &   \textbf{79.1±0.6}    &   78.9±0.8    \\
                 & PubMed       &   86.9±0.3             &   \textbf{87.0±0.3}    \\
\midrule
\multirow{3}{*}{Heterophilic} 
                 & Cornell      &   82.3±3.4             &   \textbf{83.2±4.3}    \\
                 & Texas        &   \textbf{86.9±1.5}    &   85.7±2.3    \\
                 & Wisconsin    &   84.0±2.5             &   \textbf{86.2±3.5}    \\
\bottomrule
\end{tabular}
\caption{Experiments results for comparison between low-/high- pass filters and band-stop/pass filters based on PolyGCL framework. Metrics are node classification accuracy(\%). }
\label{tab:toy_band}
\end{table}

\begin{definition}[\textbf{Spectral Signal Trend} \(\mathcal{T_\text{r}}\)]
Given node feature \(\mathbf{x}_i \in \mathbb{R}^{p}\), the spectral signal trend of node \(v_i\) across the graph can be defined as
\begin{equation}\label{eq:spectral_signal_trend}
    \mathcal{T_\text{r}}(\mathbf{x}_i)=\left(\frac{x_{i, k}}{\sqrt{d_i}}-\sum_{j \in n_i^1}\frac{x_{j, k}}{\sqrt{d_j}}\right)^2,
\end{equation}

for \(n_1 \in n_i^1\) stands for the 1-hop neighbors of node \(i\).
\end{definition}
The \textit{Spectral Signal Trend} characterizes the changes in node features on the graph, compared with node features, it introduces the additional topological message. Further, since the relative differences of \(\mathcal{T_\text{r}}(\mathbf{x})\) between nodes are only dependent on the original nodes feature, it meets our first requirement.
%TODO: Proof
Secondly, to eliminate the influence of the differences between low-pass and high-pass views and propose a composite feature(Eq.\ref{eq:genereal_C}), where \(\Delta\mathbf{x}_i\) denotes aggregated original feature differences between node \(i\) and its neighbors. We consider the two simplest composition ways, subtraction or addition of \(\mathcal{T_\text{r}}(\mathbf{x})\) between low-pass and high-pass views, where \(\Delta g(\mathbf{\Lambda})\) is equal to \((g^l(\mathbf{\Lambda})-g^h(\mathbf{\Lambda}))\) and \((g^l(\mathbf{\Lambda})+g^h(\mathbf{\Lambda}))\) respectively. However, which one is more stable?

\begin{equation}
    \mathcal{C}(\mathbf{x}_i^l,\mathbf{x}_i^h) = \left | \mathbf{U} \Delta g(\mathbf{\Lambda})\mathbf{U}^T \Delta \mathbf{x}_i \right|.
    \label{eq:genereal_C}
\end{equation}

\begin{figure*}[t]
\centering
\includegraphics[width=\textwidth]{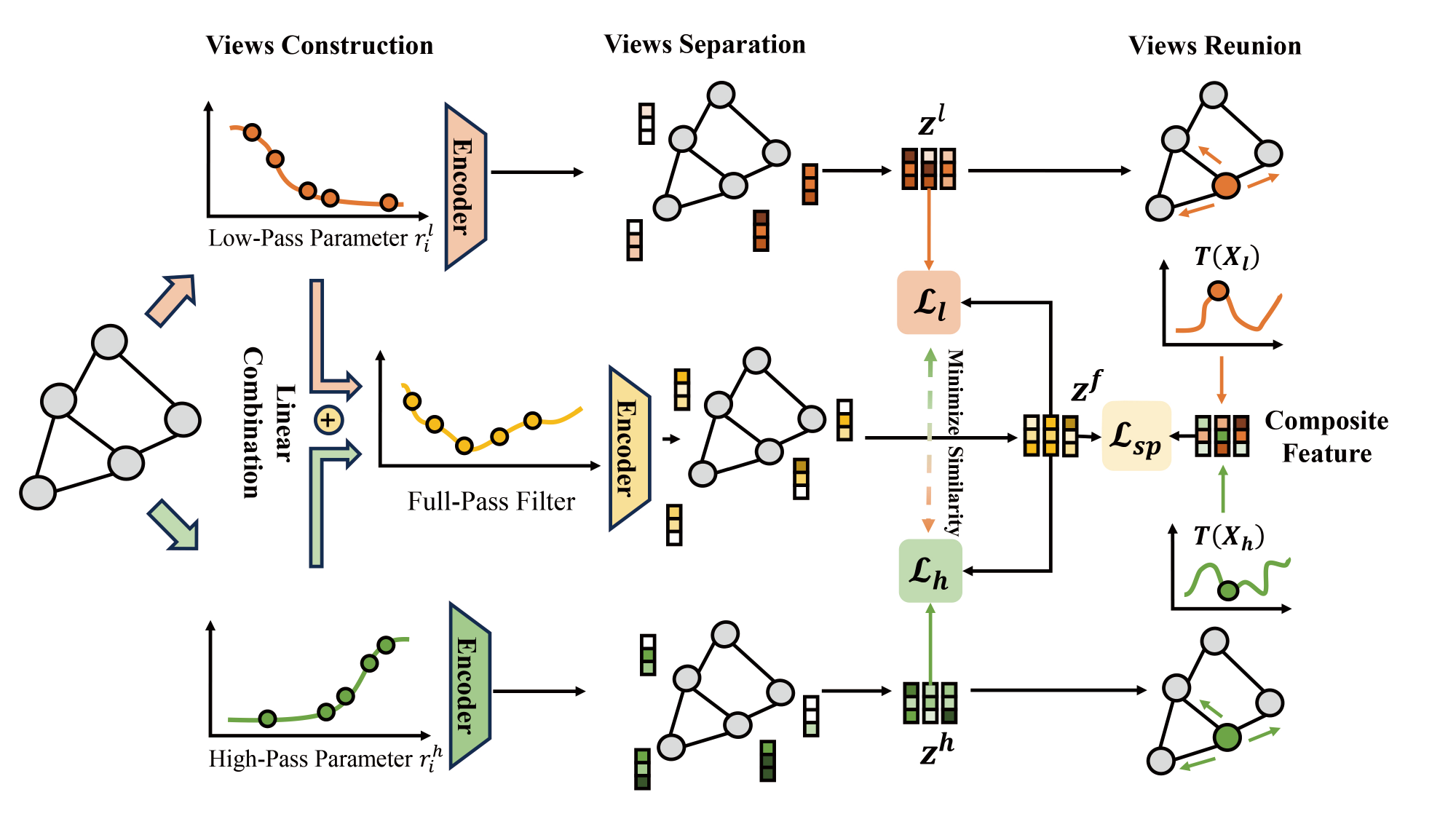}
\caption{Illustration for the overall Pipeline of LOHA. Based on the natural opposite specialties between low-pass view and high-pass view, we minimize the mutual information between these two sides to construct contrastive views and separate sub-view features to learn better filters. Further, in the views reunion section, we propose to use a composite feature based on the spectral signal trends of the two views to reunite node features. }
\label{fig:pipeline}
\end{figure*}

To better conduct theoretical analysis, derived from ~\cite{necessity}, we make the following assumptions: consider a graph \(\mathcal{G}\), where each node \(i\) has features \(\mathbf{x}_i \in \mathbf{R}^f\). Denote a composition function \(\mathcal{C}(\mathbf{x}_i^l,\mathbf{x}_i^h)\) between low-pass and high-pass views. We assume that (1) \(\mathcal{C}\) for all nodes are sampled from the same distribution; (2)  \(\mathcal{C}_i\) is independent to across each dimension; (3) The features in \(\mathbf{X}\) are bounded by a positive scalar \(B\), thus \(\max _{i, j}|\mathbf{X}[i, j]| \leq B\). Suppose that given graph filtering function \(\mathbf{U} g(\boldsymbol{\Lambda}) \mathbf{U}^\top\mathbf{X}\), the bound of transformed feature \(\mathbf{X}'\) is \(\max _{i, j}|\mathbf{X'}[i, j]| \leq \max(\lambda') \cdot B\), where \(\max(\lambda')\) denotes the maximum eigenvalue of \(\mathbf{U} g(\mathbf{\Lambda}) \mathbf{U}^T\). Thus, we have:
\begin{theorem}\label{the:prob_for_c}
%TODO: assumptions, dependency
Given the expectation of \(\mathcal{C}_i\) as \(\mathbb{E}(\mathcal{C}_i\)), \(d_i\) as the degree of node \(i\). For any \(t > 0\), the probability that the distance between \(\mathcal{C}_i\) and its expectation is larger than \(t\) is bounded by:
\begin{equation}\label{eq:hoeffdings_ours}
\fontsize{9}{12}
\mathbb{P}\left(\left\|\mathcal{C}_i-\mathbb{E}\left[\mathcal{C}_i\right]\right\|_2 \geq t\right) 
 \leq 2 \cdot f \cdot \exp \left(-\frac{d_{i} t^2}{2 (\max(\lambda'))^4 B^2 f}\right).
\end{equation}
\end{theorem}
Detailed proofs are provided in Appendix. From Theorem~\ref{the:prob_for_c}, smaller \(\max(\lambda')\) lead to more stable \(C_i\), which meets our requirements. Thus, we subtract low-pass and high-pass views to form the composite features.  

\section{Method}
The overall pipeline of LOHA is illustrated in Fig.~\ref{fig:pipeline}. In this section, we will provide a detailed explanation of our proposed method, LOHA, from the two main parts, Views Construction and Loss Design. In the Loss Design section, we separately introduce our loss design based on Views Separation and Views Reunion parts to show our insight more clearly. 
\subsection{Views Construction}
Following~\cite{chebnetii}, we adopt Chebyshev polynomials with interpolation as base polynomials, which can be formulated as \(\sum_{k=0}^K w_k T_k(\hat{\mathbf{L}}) \mathbf{X}\), where \(\hat{\mathbf{L}}=2 \tilde{\mathbf{L}} / \lambda_{\text {max }}-\mathbf{I}\), and \(w_k\) is reparameterized as follows:
\begin{equation}\label{equ:w_in_cheb}
w_k=\frac{2}{K+1} \sum_{j=0}^K \gamma_j T_k\left(x_j\right),
\end{equation}
where \(x_j=\cos \left(\frac{j+1 / 2}{K+1} \pi\right), \left(j=0, \ldots, K\right)\) denote the Chebyshev nodes for \(T_{K+1}\). As for the learnable parameter \(\gamma_j\), we propose to use a learnable sliding cosine-parametric-based formulation to build our low-pass and high-pass filters based on parameters \(\gamma_j^l\) and \(\gamma_j^h\) as follows:
\begin{equation}
%\fontsize{8}{12}
    \begin{aligned}
    \gamma_j^l&=\sigma\left(\beta_a^l\right)-\frac{1}{2} \sigma\left(\beta_b^l\right)(1+\cos (1+\frac{\frac{1}{2}\tanh(\delta_l)+j}{K} \pi)) ,\\
    \gamma_j^h&=\sigma\left(\beta_a^h\right)+\frac{1}{2} \sigma\left(\beta_b^h\right)(1+\cos (1+\frac{\frac{1}{2}\tanh(\delta_h)+j}{K} \pi)) ,
    \end{aligned}
\label{equ:gamma}
\end{equation}
where \(\sigma=ReLU(\cdot)\), \(\beta_a^l\) and \(\beta_a^h\) are initialized to 0 and 2 respectively whereas \(\beta_b^l\) and \(\beta_b^h\) are set to 2 and these four parameters are all trainable during the learning process. Noted that compared with filters proposed by~\cite{s3gcl}, we add the learnable \(\tanh(\delta)\) parameters on \(j\) to achieve our ``sliding'' propose, which is a simple trick but shows powerful capability. All \(\delta\) are initialized to \(0\) and \(\frac{1}{2}\tanh(\cdot)\) function constrain the values to lie inside \(\left(-0.5, 0.5\right )\), which ensures the \(\gamma_i\) will not intersect and strictly obey the low-pass or high-pass sequences accordingly. As stated in~\cite{s3gcl}, when the cosine-parameterized strategy effectively emphasizes key frequencies while reducing less important ones, compared with traditional equidistant interpolation, we believe that adding a sliding strategy could give the filters more flexibility. The capability of this strategy is validated in ablation experiments. To sum up, substituting \(\gamma_i\) with \(\gamma^l\) and \(\gamma^h\) in Eq.\ref{equ:w_in_cheb} respectively, we could obtain two distinct views as in Eq.\ref{eq:encoder}.
\begin{equation}\label{eq:encoder}
\fontsize{8}{12} 
\mathbf{Z}_l=MLP\left(\sum_{k=0}^K w_k^l T_k(\tilde{\mathbf{L}}) \mathbf{X}\right),\mathbf{Z}_h=MLP\left(\sum_{k=0}^K w_k^h T_k(\tilde{\mathbf{L}}) \mathbf{X}\right),
\end{equation}
where \(MLP\) are shared between two views. Given feature \(\mathbf{Z}_l\) and \(\mathbf{Z}_h\), we obtain the final embedding through linear combination \(\mathbf{Z}_f = \alpha \mathbf{Z}_l + \beta \mathbf{Z}_h\). Next, we introduce our designed loss for contrastive learning.

\subsection{Loss Design}
Our loss function consists of three parts:
\begin{equation}\label{equ:total_loss}
    \mathcal{L} = \mathcal{L}_l + \mathcal{L}_h + \mu\mathcal{L}_{sf},
\end{equation}
for which \(\mathcal{L}_l\) and \(\mathcal{L}_h\) belong to views separation part to push node features in low-pass view and high-pass view apart and \(\mathcal{L}_{sf}\) as views reunion part to reunite two views of one node to be aggregated. Hyper-parameter \(\mu\) is to balance \(\mathcal{L}_{sf}\) with \(\mathcal{L}_l\) and \(\mathcal{L}_h\).
\subsubsection{Views Separation.}
Inspired by InfoNCE loss~\cite{infoNCE}, we formulate our \(\mathcal{L}_l\) and \(\mathcal{L}_h\) as follows:
\begin{equation}
\fontsize{8}{12} 
\begin{aligned}
& \mathcal{L}_l=-\frac{1}{|\mathcal{V}|} \sum_{v_i \in V}\left(\log \frac{s\left(z_i^{f}, z_i^{l}\right)}{\sum_{v_p \in \mathcal{V} \setminus v_i} s\left(z_i^{f}, z_p^l\right)+s\left(z_i^l, z_i^h\right)}\right), \\
& \mathcal{L}_h=-\frac{1}{|\mathcal{V}|}\sum_{v_i \in V}\left(\log \frac{s\left(z_i^{f}, z_i^h\right)}{\sum_{v_p \in \mathcal{V} \setminus v_i} s\left(z_i^f, z_p^h\right)+s\left(z_i^l, z_i^h\right)}\right). 
\end{aligned}
\label{equ:loss_lh}
\end{equation}
%TODO: setminus or not
%where \(n^1_i\) represents the set of one-hop neighbors and itself for node \(v_i\). 
Here, we define \(s(z_i^{f}, z_p^l) = \exp(\text{sim}(z_i^f, z_p^l) / \tau)\), where \(z_i^f\), \(z_i^l\), and \(z_i^h\) are embeddings from the final full-pass, low-pass, and high-pass filters respectively, and \(\tau\) is the contrast temperature. To enhance the training of the final full-pass filters, we consider the features obtained from the full-pass view \(z_i^f\) and the low-/high-pass views (\(z_i^l\)/\(z_i^h\)) of a node \(v_i\) as the positive pair in the design of \(\mathcal{L}_l\) and \(\mathcal{L}_h\). Furthermore, we treat features of other nodes \(v_p\) from the corresponding low-/high-pass (\(z_p^l\)/\(z_p^h\)) as negative samples of \(z_i^f\). To enforce the low-pass view to learn feature smoothness and the high-pass view to highlight neighborhood differences, we treat \(z_i^l\) and \(z_i^h\) of node \(v_i\) as negative samples of each other. This approach leverages the distinct properties of these two views effectively.

\subsubsection{Views Reunion.}
As we discussed before, when \(z_i^l\) and \(z_i^h\) are both features for node \(v_i\) and we minimize their mutual information in the views separation part, we propose to use the differentiate of spectral signal trends of nodes to reunite node features. Our \(\mathcal{L}_{sf}\) are determined:
\begin{equation}\label{equ:L_sf}
\fontsize{8}{12} 
\begin{aligned}
&\mathcal{L}_{sf} = -\frac{1}{|\mathcal{V}|} \sum_{v_i \in V}\left(\log \frac{s\left(z_i^{f}, \mathcal{C}^-(z_i^l,z_i^h)\right)}{\sum_{v_p \in \mathcal{V} \setminus v_i} s\left(z_i^{f}, \mathcal{C}^-(z_p^l,z_p^h)\right)}\right), \\
&where \quad \mathcal{C}^-(\mathbf{x}_i^l, \mathbf{x}_i^h) =\left|\mathbf{U}\left(g^l(\mathbf{\Lambda})-g^h(\mathbf{\Lambda})\right)\mathbf{U}^T\Delta\mathbf{x}_i \right|.\\
\end{aligned}
\end{equation}
Noted that here we define positive and negative pairs in a relative concept, for which we reckon that the similarity between features of the node \(v_i\) from full-filter views \(z_i^f\) and spectral differences \(\mathcal{C}^-(z^l, z^h)\) should be higher than that with other nodes \(v_p\). 

What's more, to further validate the effectiveness of Eq.\ref{eq:spectral_signal_trend}, we propose two variants of it when computing \(\mathcal{C}^-(\mathbf{x}^l_i, \mathbf{x}^h_i)\). Specifically, by changing the input \(\mathbf{x}_i\), we have:
\begin{equation}
\begin{aligned}
     &\mathcal{T_\text{r}}_1(\hat{\mathbf{x}_i})=\left(\hat{\mathbf{x}_i}-\sum_{j \in n_i^1}\hat{\mathbf{x}_j}\right)^2 \\
     where &\quad \hat{\mathbf{x}_i} = \frac{\tilde{\mathbf{x}_i}}{\sum_i \tilde{\mathbf{x}_i}} \quad and  \quad \tilde{\mathbf{x}_i}={\sum_k\frac{\mathbf{x}_{i,k}}{\sqrt{d_i}}},
\end{aligned}
\label{eq:trend_sum}
\end{equation}
     
\begin{equation}
\begin{aligned}
     &\mathcal{T_\text{r}}_3(\widehat{\mathbf{x}_i}) = \left( \widehat{\mathbf{x}_i} - \sum_j\widehat{\mathbf{x}_j}     \right)^2 \\
     where \quad &\widehat{\mathbf{x}_i}=  \left[ \hat{\mathbf{x}_i} \| \mathrm{mean}\left(\frac{\mathbf{x}_{i,k}}{\sqrt{d_i}}\right) \| \mathrm{std}\left(\frac{\mathbf{x}_{i,k}}{\sqrt{d_i}}\right) \right],
\end{aligned}
\label{eq:trend_three}
\end{equation}
where '\(||\)' means feature concatenation. Thus, the feature is squeezed from \(\mathcal{N}\times\mathcal{D}_{hidden}\) to \(\mathcal{N}\times 1\) for Eq.\ref{eq:trend_sum} and \(\mathcal{N}\times 3\) for Eq.\ref{eq:trend_three}, thus holds different information density. 

\begin{table*}[t]
\centering
\fontsize{9}{12} 
\setlength{\tabcolsep}{1mm}
\begin{tabular}{lccccccccccc}
\toprule
& Methods         & Cornell  & Texas    & Wisconsin  & Actor   & Chameleon & Amazon     & Cora       & Citeseer   & PubMed   &Avg.  \\ \midrule
\multicolumn{11}{c}{Full-Supervised} \\ \midrule
 & MLP            & 86.4±2.1 & 87.0±2.3 & 86.3±3.8* & 40.2±0.5 & 46.1±1.2  & 39.1±1.7   & 76.5±0.8 & 74.3±1.0 & 85.9±0.4     &69.1\\
 & GCN            & 68.6±3.8 & 78.2±3.1 & 74.5±3.1 & 34.2±0.9 & 58.8±2.4  & 43.3±0.4   & 85.3±1.0 & 76.7±0.5 & 85.7±0.2      &67.3\\
 & ChebNet        & 83.6±7.5 & 85.3±2.5 & 82.5±4.9 & 37.6±1.7 & 59.7±2.4  & 47.8±0.7*   & 85.5±1.2 & 77.0±0.7 & 86.7±0.4     &71.7\\
 & BernNet        & 86.7±0.4 & 87.2±1.5 & 84.4±1.8 & 41.8±1.3* & 69.3±1.5  & 44.3±0.1   & 87.4±0.9* & 78.0±0.9* & 86.5±0.2   &74.0\\
 & ChebNetII      & 88.4±1.6*& 89.3±2.3* & 85.6±2.5 & 40.1±0.8 & 71.5±1.2*  & 45.1±0.5   & 86.2±1.1 & 76.5±0.9 & 87.3±0.3*   &74.4*\\ \midrule
\multicolumn{11}{c}{Self-Supervised} \\ \midrule
 & DGI            & 72.4±6.9 & 81.5±3.2 & 76.4±1.9 & 32.6±1.2 & 58.2±0.8  & 42.8±0.4   & 84.9±0.9 & 76.4±0.9 & 83.1±0.6      &67.6\\
 & GraphCL        & 61.5±5.7 & 68.0±5.0 & 61.5±3.9 & 32.8±1.3 & 59.2±1.4  & 39.0±0.3   & \underline{86.4±0.7} & 78.0±0.5 & 85.2±0.2 &63.5\\
 & GRACE          & 60.7±9.8 & 75.6±2.8 & 72.7±3.2 & 31.9±1.0 & 59.6±1.3  & 43.3±0.2   & 83.5±0.7 & 74.2±0.8 & 81.8±0.6             &64.8\\
 & BGRL           & 59.8±3.0 & 70.7±2.8 & 63.1±4.0 & 33.1±0.7 & 64.0±1.3  & 41.6±0.5   & 84.3±0.7 & 74.5±1.0 & 83.2±0.3             &63.8 \\
 & GREET          & 78.4±3.9 & 80.0±3.9 & 83.6±3.7 & 38.3±0.9 & 63.2±1.0  & 42.0±0.2   & 85.7±1.4  & \underline{78.0±0.4} & 85.6±0.3 &70.5\\
 & GraphACL       & 60.2±0.5 & 70.8±0.7 & 73.6±0.5 & 29.6±0.1 & 72.2±2.3  & 41.4±0.6   & 86.3±1.0 & 77.4±0.7 & 85.8±0.3               &66.4\\
 & PolyGCL        & 82.4±3.4 & 87.9±1.5 & 84.0±2.5 & \underline{41.8±0.8} & \underline{71.9±1.2}  & 47.3±0.4   & \textbf{87.6±0.7*} & \textbf{78.9±0.6*} & \underline{86.9±0.3} &74.3   \\ \cmidrule(lr){2-12}
 & \textbf{LOHA}  & \textbf{90.7±1.3*} & \textbf{93.4±1.3*} & \textbf{95.9±0.9*} & \textbf{42.8±0.8*} & \textbf{73.0±0.9*}  & \textbf{48.8±0.3*}&85.1±1.3  &77.1±0.5 &\textbf{87.3±0.3*}    &\textbf{77.1*}\\
 & LOHA-var-3     & \underline{89.0±2.3} & \underline{91.4±1.5} &94.3±1.5  & 40.5±0.8  &70.2±0.8  &\underline{48.0±0.5}   &83.1±0.9    &76.2±0.3    &86.8±0.8            &\underline{75.5}  \\
 & LOHA-var-1     & 88.4±2.1 & 89.3±1.6 &\underline{95.5±1.4}  & 38.7±1.2  &68.4±1.0  &47.3±0.8   &83.5±1.0    &74.9±0.8    &86.3±1.3            & 74.7 \\
                \bottomrule 
\end{tabular}
\caption{Mean node classification accuracy (\%) on real-world graph. The best full-supervised results are marked by *. For self-supervised models, \textbf{boldface} letters indicate the best results and \underline{underlining} letters denote the runner-up results. The best results that surpass full-supervised performance are also marked with *. LOHA-var-3 and LOHA-var-1 are variants correspond to models with Eq.\ref{eq:trend_three} and~\ref{eq:trend_sum} respectively. }
\label{tab:main_exp}
\end{table*}

\section{Experiment}
In this section, we present a comprehensive evaluation of LOHA, through a series of node classification experiments on 9 real-world datasets. Comparison between LOHA with other baselines and ablation studies validate the effectiveness of LOHA and help us to gain further insights. 

\subsection{Baselines and Settings}
To show the superiority of LOHA, we select several common and typical full-supervised networks and novel self-supervised graph contrastive learning frameworks as our baseline as follows: (1)Full-Supervised Networks: MLP, GCN~\shortcite{GCN}, ChebNet~\shortcite{chebnet}, BernNet~\shortcite{bernnet}, and ChebNetII~\shortcite{chebnetii}; (2) Self-supervised Graph Contrastive Learning Frameworks: DGI~\cite{dgi}, GraphCL~\cite{GraphCL}, GRACE~\shortcite{grace}, BGRL~\cite{BGRL}, GREET~\shortcite{greet}, GraphACL~\cite{GraphACL} and PolyGCL~\shortcite{polygcl}.
\subsubsection{Evaluation Protocol.}
Following~\cite{dgi}, we use the widely adopted two-stage linear evaluation pipeline in graph contrastive learning. In the first stage, the node features and structure information are self-trained without any label information. After convergence, the output feature of the first stage would then be fixed and used to train, validate, and test through a simple classifier. We follow the training and validation strategies as ~\cite{gpr-gnn}, where nodes are randomly split into 60\%, 20\%, and 20\%. All comparative methods share the same fixed random splits. Output embedding size and hyper-parameters in stage 2 are also fixed for fair comparison. More detailed settings can be found in Appendix. 
%TODO: experiment settings in appendix

\subsection{Datasets}
We choose widely used real-world datasets with different homophily levels to evaluate the performance of LOHA. (1) Homophilic Graphs: Cora, Citeseer, and PubMed from ~\cite{cora-dataset}. (2) Heterophilic Graphs: Cornell, Texas, Actor and Wisconsin from~\cite{geom-gcn};  Chameleon from ~\cite{squirrel-dataset}; Amazon-ratings(Amazon for short in tables) from~\cite{amazon_ratings}. Details of the dataset can be found in Appendix. 
%TODO: Dataset intro in appendix

\subsection{Results}
The results, as presented in Table~\ref{tab:main_exp}
, demonstrate that LOHA consistently achieves strong performance across all datasets, with particularly notable success on heterophilic graphs. Specifically, LOHA surpasses all baseline models on 7 out of 9 benchmarks, indicating its robustness and adaptability across different graph structures. Notably, on the heterophilic graphs \textit{Cornell}, \textit{Texas}, and \textit{Wisconsin}, LOHA exhibits substantial relative improvements over the runner-up models, achieving gains of 8.3\%, 5.5\%, and 11.9\% respectively. These improvements highlight LOHA's ability to effectively capture and leverage the structural nuances inherent in graphs with lower homophily levels.

Moreover, LOHA not only outperforms self-supervised baselines but also surpasses fully-supervised models on all heterophilic datasets, underscoring its capacity to generalize well in scenarios where node labels are limited or unavailable. This is particularly significant considering that fully-supervised methods typically have access to ground-truth labels during training. LOHA variants further reinforce the framework’s effectiveness, consistently delivering superior performance across various settings. For example, LOHA-var-3 outperforms the next best self-supervised model by an average of 1.2\% across all datasets, demonstrating the robustness of the proposed approach. Similarly, LOHA-var-1 also shows competitive results.

It is noteworthy that the baseline model, PolyGCL, utilizes a node shuffling strategy to increase the number of negative samples, which is actually a kind of data augmentation. In contrast, our LOHA is an 'augmentation-free' method that simply leverages the intrinsic differences between low-pass and high-pass views, but still shows incredible improvements. To more intuitively verify the effectiveness of the proposed loss, we replace the original loss function of PolyGCL with LOHA ones. The results, presented in Table~\ref{tab:ab_poly_loss}, demonstrate the competitive performance of our approach and the advantages gained through the direct contradiction between low-pass and high-pass views.

\begin{table}[t]
\setlength{\tabcolsep}{1mm}
\centering
\begin{tabular}{clll}
\toprule
                {Datasets}   &PolyGCL                   & Poly-LOHA             &Poly-LOHA-3\\
\midrule
                 Cora         &\textbf{87.57±0.73}      &86.21±0.65             &83.32±0.13\\ 
                 Citeseer     &\textbf{79.13±0.58}      &78.05±0.91             &75.21±0.38\\%vector1
                 PubMed       &86.91±0.34               &\textbf{86.96±0.32}    &85.45±0.41\\%vector1
\midrule
                 Cornell      &82.36±3.42               &\textbf{86.73±2.34}    &85.12±3.32\\ 
                 Texas        &87.92±1.46               &\textbf{91.51±0.92}    &90.36±1.45\\  
                 Wisconsin    &84.03±2.52               &\textbf{91.06±1.64}    &88.52±1.98\\ 
                 Actor        &41.76±0.75               &\textbf{43.12±0.63}    &42.08±0.83\\ %vector1
                 Chameleon    &71.92±1.16               &\textbf{73.25±1.14}    &72.04±1.35\\ 
                 Amazon       &47.31±0.41               &\textbf{48.61±0.24}    &46.38±0.51\\
\bottomrule
\end{tabular}
\caption{Experiments results(Node Classification Accuracy(\%)) for ablation study between origin PolyGCL and the variant of PolyGCL which substitutes the loss function with LOHA loss(Poly-LOHA) and LOHA loss variant with Eq.\ref{eq:trend_three} (Poly-LOHA-3).}
\label{tab:ab_poly_loss}
\end{table}

\subsubsection{Variants of LOHA.}
\begin{figure}[t] 
\centering
\includegraphics[width=0.8\columnwidth]{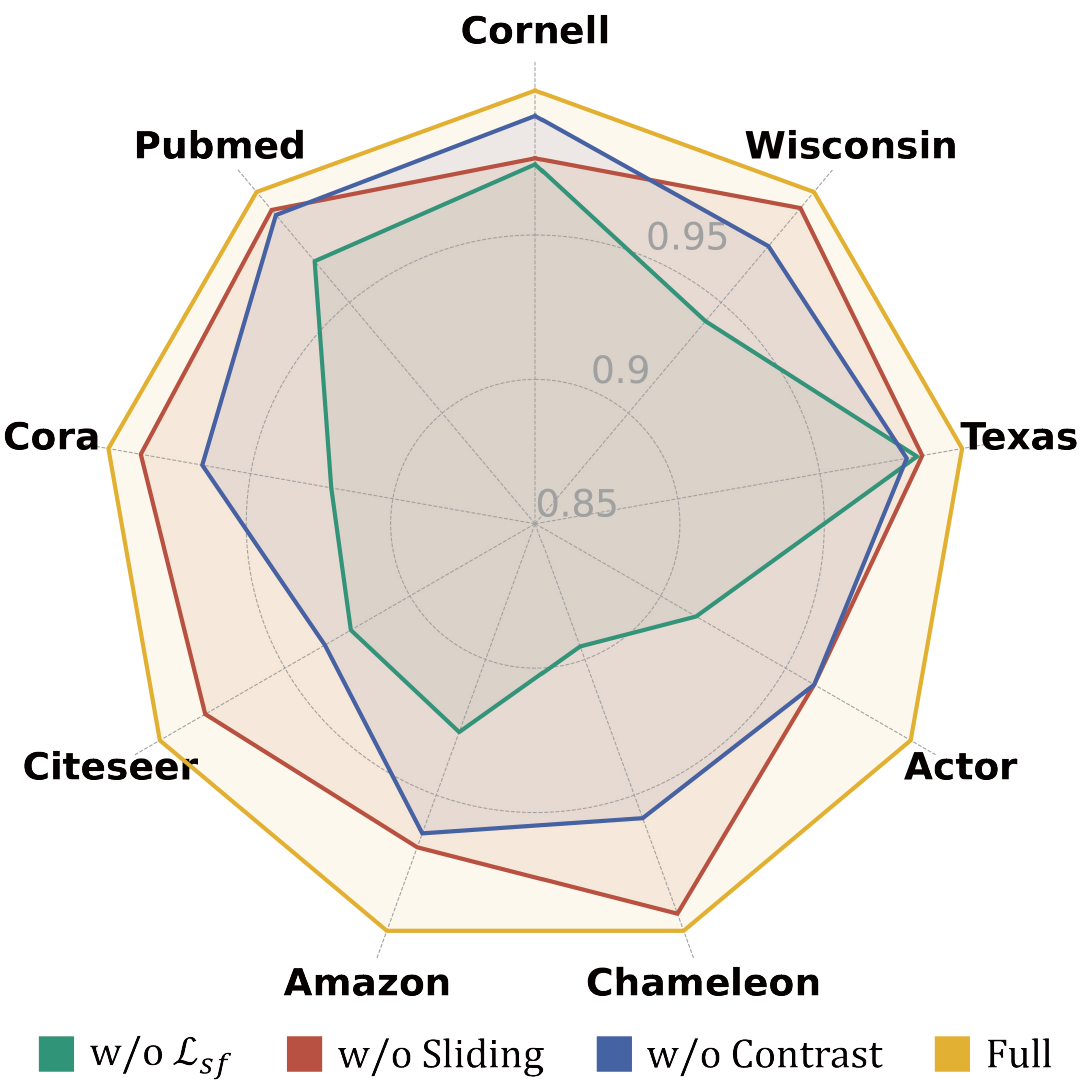}
\caption{Experiments results for ablation study between LOHA and its variants. In order to better demonstrate the effect, we used the results of the full model as the denominator for each result for normalization.}
\label{fig:abl_all}
\end{figure}

To analyze the effect of each part, we introduce three variants of LOHA:(1)w/o sliding: deleting the sliding trick in Eq.\ref{equ:gamma};(2)w/o \(\mathcal{L}_{sp}\): deleting the \(\mathcal{L}_{sp}\) in Eq.\ref{equ:L_sf}; (3)w/o contrast: deleting \(s\left(z_i^l, z_i^h\right)\) in Eq.\ref{equ:loss_lh}. The results are shown in Fig.~\ref{fig:abl_all}, which reveal that the full model consistently achieves the best performance across all datasets. The experimental results demonstrate the effectiveness of each of these three parts. Particularly, the absence of \(\mathcal{L}_{sp}\) leads to considerable performance degradation, especially on homophilic graphs like \textit{Cora}. It indicates that \(\mathcal{L}_{sp}\) plays a crucial role in maintaining the model's performance on graphs where feature smoothness is more prevalent. Similarly, when the contrastive term is excluded, this variant shows the largest performance drop across all datasets, underscoring its importance in capturing the essential differences between low-pass and high-pass views. Although sliding interpolation is just a minor trick, it also displays certain advantages. Overall, the results suggest that while each component contributes to LOHA's effectiveness, their combination is particularly powerful, achieving optimal results by balancing the strengths of each.

\section{Related Work}
\subsection{Spectral Graph Neural Network}
As a low-pass filter, the suboptimal performance of traditional GCN on heterophilic graph simulated the exploration and design of complex SGNN. ~\cite{chebnet} introduces ChebNet, which employs Chebyshev polynomials to achieve localized filters with reduced computational complexity, making it suitable for large-scale graphs. Furthermore, ~\cite{chebnetii} enhances the original ChebNet by eliminating the Runge phenomenon. BernNet~\shortcite{bernnet} leverages the flexibility of Bernstein polynomials to design adaptive spectral filters that can better handle heterophilic graphs by learning complex filter shapes. Similarly, ASWT-SGNN~\shortcite{aswt-sgnn} uses flexible spectral wavelet transforms to capture multiscale features and enhance graph representation learning under self-supervision. However, when these methods work well in the supervised domain, how to obtain a set of optimal filters automatically and better utilize the advantages of both low-frequency and high-frequency information in self-supervised workflow remains uncovered.  

\subsection{Graph Contrastive Learning}
Graph contrastive learning aims to learn consistent representations from different views to mine invariant information of graph data.  Through experimental discovery, SPAN ~\shortcite{span} points out that the insight of augmented GCL is to maximize spectral changes, thus proposing an augmentation method to maximize the spectral differences between two views. However, SpCo~\shortcite{spco} deems that the inner key lies in the difference between high-frequency and low-frequency parts, which the former one should be higher than the counterpart.
Without augmentation, some methods contrast opposing views by inputting the same data into different encoders and pushing together the representations of the same entities from different views. For example, PolyGCL\shortcite{polygcl} utilizes polynomial filters to construct low-pass and high-pass views and combine them through linear addiction. Similarly, S3GCL\shortcite{s3gcl} uses MLP to better correlate the results of two views thus reducing inference time. However, these two methods lack a direct comparison between the two subviews, which raises doubts about whether they truly and effectively apply the advantages of low-frequency and high-frequency information as expected. Thus, unlike the traditional methods, we propose to minimize the similarities between low-pass and high-pass views to force the model to maximize their specialties thus obtain better features. 

\section{Conclusion}
This paper introduces LOHA, a novel graph contrastive learning (GCL) framework that leverages the inherent opposition between low-pass and high-pass views. By focusing on maximizing the distinction between these views, LOHA enhances filter effectiveness and improves representation learning. Recognizing that excessive differentiation may cause node feature separation, we further introduce a high-dimensional feature, spectral signal trends, to serve as an extra feature that encourages feature reunion for individual nodes. In essence, LOHA is a simple yet powerful framework that optimizes filter learning through contrastive views and subsequently reunites node features using spectral signal trends. Extensive experiments and ablation studies demonstrate LOHA's superiority across both homophilic and heterophilic graphs.

\section*{Acknowledgments}
This work was supported by the National Natural Science Foundation of China (grant nos. 62072384 and 62372391) and Fujian Provincial Major Science and Technology Project(2022YZ040011).

\bibliography{aaai25}

\end{document}